\documentclass[review]{elsarticle}
\graphicspath{ {./figures/} }
\usepackage{hyperref}
\usepackage{float}
\usepackage{verbatim} 
\usepackage{apalike}
\usepackage{amssymb}
\usepackage{amsmath,amssymb,amsfonts}
\usepackage{mathtools,cancel}
\restylefloat{figure}
\restylefloat{table}

\biboptions{authoryear}

\begin{document}
\newcommand{\bigCI}{\mathrel{\text{\scalebox{1.07}{$\perp\mkern-10mu\perp$}}}}
\newcommand{\nbigCI}{\cancel{\mathrel{\text{\scalebox{1.07}{$\perp\mkern-10mu\perp$}}}}}
\begin{frontmatter}
\title{Deep Recurrent Modelling of Granger Causality with Latent Confounding}

\author[label1]{Zexuan Yin\corref{cor1}} 
\ead{zexuan.yin.20@ucl.ac.uk}


\author[label1]{Paolo Barucca}
\ead{p.barucca@ucl.ac.uk}

\cortext[cor1]{Corresponding author.}
\address[label1]{Department of Computer Science, University College London, WC1E 7JE, United Kingdom}

\begin{abstract}
Inferring causal relationships in observational time series data is an important task when interventions cannot be performed. Granger causality is a popular framework to infer potential causal mechanisms between different time series. The original definition of Granger causality is restricted to linear processes and leads to spurious conclusions in the presence of a latent confounder. In this work, we harness the expressive power of recurrent neural networks and propose a deep learning-based approach to model non-linear Granger causality by directly accounting for latent confounders. Our approach leverages multiple recurrent neural networks to parameterise predictive distributions and we propose the novel use of a dual-decoder setup to conduct the Granger tests. We demonstrate the model performance on non-linear stochastic time series for which the latent confounder influences the cause and effect with different time lags; results show the effectiveness of our model compared to existing benchmarks.
\end{abstract}

\begin{keyword}
 latent confounders \sep recurrent neural networks \sep time series prediction
\end{keyword}

\end{frontmatter}

\section{INTRODUCTION}
\label{introduction}
Identifying causal relationships from time series data is important as it helps to facilitate informed decision making. When controlled experiments are feasible, interventions are often performed to break the symmetry of association and provide the direction of causal mechanisms \citep{Eichler2012}, e.g. predicting patients' response to certain treatments over time \citep{Bica2020}. In reality, causal inference through interventions is not always feasible as it could be unethical, costly, or simply impossible to carry out, as in the case of financial time series \citep{Hiemstra1994} and climate variables \citep{Stips2016}; in those scenarios we resort to causal inference from observational data. 

Granger causality \citep{Granger1969} is a commonly used framework to infer potential causal relationships. The notion of Granger causality relies on two fundamental principles: 1. the cause precedes the effect in time and 2. the cause contains unique information about the effect not available elsewhere \citep{Eichler2012}. We say that one time series Granger causes another if its past helps to predict the future values of the target time series \citep{Granger1969}. Traditionally, model-based Granger causality has been tested mostly on linear dynamics in the form of a vector autoregressive model (VAR) \citep{Yuan2020}, where one regresses the lagged values of potential causes against the future value of the target series and assess whether the coefficients are statistically different from zero. 

Since real world temporal dynamics are rarely linear, several adaptations to model nonlinear causal relationships have been made using for example polynomial autoregression models \citep{Bezruchko_2008} and kernel-based methods \citep{Marinazzo2011}. Model-free approaches such as transfer entropy \citep{Vicente2011} are able to detect nonlinear dependencies between time series, however they suffer from high variance and require large amounts of data for reliable estimation \citep{Tank2021}. In this work, we follow a recent trend that uses neural networks to infer complex nonlinear causal dependencies in time series data \citep{Khanna2020,Nauta2019,Tank2021,Bussmann2020,Trifunov2019,DeBrouwer2020,Marcinkevics2021,Moraffah2021}.

An important consideration for causal inference from observational time series is confounding bias. A confounder variable affects both cause and effect and therefore must be accounted for to avoid spurious conclusions. Granger causality relies on the causal sufficiency (no latent confounding) assumption \citep{Spirtes2016} and is known to be biased in the presence of confounding \citep{10.5555/3202377}. Consider the case where the confounder $Z$ affects the cause variable $X$ with lag 2 and the effect variable $Y$ with lag 4, assuming causal sufficiency would lead to the biased conclusion that $X$ Granger causes $Y$. When all confounders are observed, one could apply multivariate conditional Granger causality  tests \citep{Chen2006}, which relies on the fact that all variables that could have had a possible influence have been considered in the analysis \citep{Marinazzo2011}. In reality, it is rarely possible to measure all the confounders, nevertheless, we may have access to noisy measurements of proxies for the confounders \citep{Louizos2017,Pearl2010}. Since the majority of existing works on neural network-based approaches to Granger causality assume causal sufficiency, how best to account for latent confounders is still an open question. In this work, we use recurrent neural networks (RNN) to infer representations of the latent confounder from the available proxies, which we then use in the subsequent Granger causality tests. 

\begin{figure}[h]
    \centering
	\includegraphics[width=0.3\columnwidth,clip,keepaspectratio]{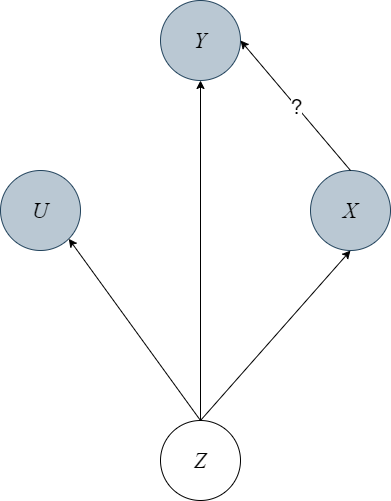}
	\caption{Causal graph showing the relationship between effect variable $Y$, cause variable $X$, latent confounder $Z$ and proxy variable $U$.}
	\label{fig1}
\end{figure}

We consider the causal system in Fig. \ref{fig1} involving a cause variable $X \in \mathbb{R}^{1 \times T}$, an effect variable $Y \in \mathbb{R}^{1 \times T}$, a latent confounder $Z \in \mathbb{R}^{1 \times T}$ and proxies of the confounder $U \in \mathbb{R}^{n \times T}$, where $T$ is the length of the time series and $n$ is the number of proxies available. Our aim is infer the Granger causal relationship between the confounded pair $X$ and $Y$. The main contributions of our work are as follows:

\begin{enumerate}
    \item we propose a deep learning-based test for nonlinear Granger causality with latent confounding. We leverage the expressive power of recurrent neural networks to infer representations of the latent confounder from the proxy variables and parameterise two predictive distributions for the target variable (with and without $X$). We apply a two-sample t-test to establish whether the inclusion of $X$ results in a statistically significant reduction in prediction error, and hence a Granger causal relationship. 
    \item we propose the novel use of a dual-decoder setup corresponding to the two predictive distributions mentioned above. This avoids the need to train two separate neural networks for comparison of predictive accuracy
    \item we demonstrate the effectiveness of our model on datasets with known data generating processes and we perform a range of sensitivity analyses to show the robustness of our proposed approach 
\end{enumerate}

\section{RELATED WORK}
\label{related work}

The original definition of Granger causality \citep{Granger1969} involves linear dynamics studied using a VAR model. For a collection of $k$ time series $\boldsymbol{X} \in \mathbb{R}^{k \times T}$ and $\boldsymbol{X}_t \in \mathbb{R}^{k}$ a VAR model is defined:
\begin{equation}
    \boldsymbol{X}_t=\sum_{l=1}^{L}\boldsymbol{A}^{(l)}\boldsymbol{X}_{t-l}+\epsilon_t,
\end{equation}
where $L$ is the maximum lag considered, $\boldsymbol{A}^{(l)}$ is a $k\times k$ matrix of coefficients and $\epsilon_t$ is a noise term with zero mean. In the linear regime, time series $j$ does not Granger-cause series i if for all $l$ $\boldsymbol{A}_{ij}^{(l)}=0$. \cite{Tank2021} generalise the definition of Granger causality for nonlinear autoregressive models:
\begin{equation}
 X_{ti}=g_{i}(X_{<t1},...,X_{<tk})+\epsilon_{ti}, 
\end{equation}
where $X_{<ti}$ denotes the history of time series $i$ and $g_{i}$ is a nonlinear function mapping the lagged values of other time series to series $i$. Granger non-causality is concluded between series $i$ and $j$ if for all $(X_{<t1},...,X_{<tk})$ and all $X^{'}_{<tj} \neq X_{<tj}$, $g_{i}(X_{<t1},...X_{<tj},...,X_{tk})=g_{i}(X_{<t1},...X^{'}_{<tj},...,X_{tk})$, implying that $g_{i}$ does not depend on$X_{<tj}$.

In \cite{Tank2021} the function $g_{i}$ is parameterised by a multilayer perceptron (MLP) regularised by group lasso penalties and trained with proximal gradient descent to shrink the input weights of lagged values of non-causal time series to zero. \cite{Bussmann2020} propose a neural additive VAR model with each time series expressed as a sum of nonlinear functions of the other time series. The nonlinear functions are parameterised by MLPs and the additive structure allows the contribution of each time series to be analysed separately. 

\cite{Nauta2019} propose an attention based convolutional neural network with an explicit validation phase. The attention mechanism learns which time series are attended to during prediction, and interventions on potential causal time series are performed in the validation phase. \cite{Khanna2020} infer Granger causal relations from a structured sparse estimate of internal parameters of statistical recurrent units \citep{Oliva2017} trained for time series prediction.

A popular class of methods involves training two neural network time series prediction models and comparing their performances. One model would accept the past values of the target and exogenous variables as inputs, and the other accepts only the past target values. A statistically significant reduction in prediction error is a sign of Granger causality. In existing literature, these prediction models are often different variants of RNNs \citep{Wang2018,Duggento2019,Abbasvandi2019} or MLPs \citep{Orjuela-Canon2020}. Our proposed approach falls within this class of methods however we argue that training two separate neural networks is inefficient and spurious conclusions could be reached due to differences in neural network hyperparameters. In our architecture, we propose to train a neural network with two decoders to alleviate these issues.

With the exception of \cite{Nauta2019}, all above-mentioned literature assumes causal sufficiency. How best to account for an unobserved confounder in Granger causal analysis is an open question. In \cite{Nauta2019}, the model can only detect a latent confounder if it affects cause and effect with equal time lags. In our work however we consider a more challenging scenario involving different lags in the causal mechanisms. We follow a popular approach involving the use of neural networks to infer representations of the latent confounder (a substitute confounder). \cite{Louizos2017} propose a variational autoencoder to recover the joint distribution of the observed and latent variables which they use to estimate the average treatment effect (ATE) in a static setting. \cite{Trifunov2019} adapt the architecture in \cite{Louizos2017} to a time series setting for the estimation of ATE. \cite{Bica2020} propose a recurrent neural network architecture to build a factor model and estimate ATE using the inferred substitute confounders.  

Outside of the deep learning domain, different methods can accommodate hidden confounders to different extents. \cite{Chu2008} propose additive nonlinear time series model (ANLTSM) which can only deal with hidden confounders that are linear and instantaneous. Conditional independence based approaches LPCMCI \citep{Gerhardus2020} and SVARFCI\citep{Malinsky2018} detect hidden confounders by inferring a special edge type in the partial ancestral graph.

\section{METHODOLOGY}
\label{methodology}

Our proposed approach involves the use of multiple recurrent neural networks to parameterise predictive distributions. We define the full model predictive distribution as $P(Y_{t+1}|Y_{1:t},X_{1:t},Z_{1:t})$. The restricted model distribution is defined as $P(Y_{t+1}|Y_{1:t},Z_{1:t})$. Parameterising the two predictive distributions allows us to compare the predictive performances of two time series prediction models and a statistically significant reduction in prediction error from the restricted model to the full model is a sign of Granger causality. 

\begin{figure}[h]
    \centering
	\includegraphics[width=0.5\columnwidth,clip,keepaspectratio]{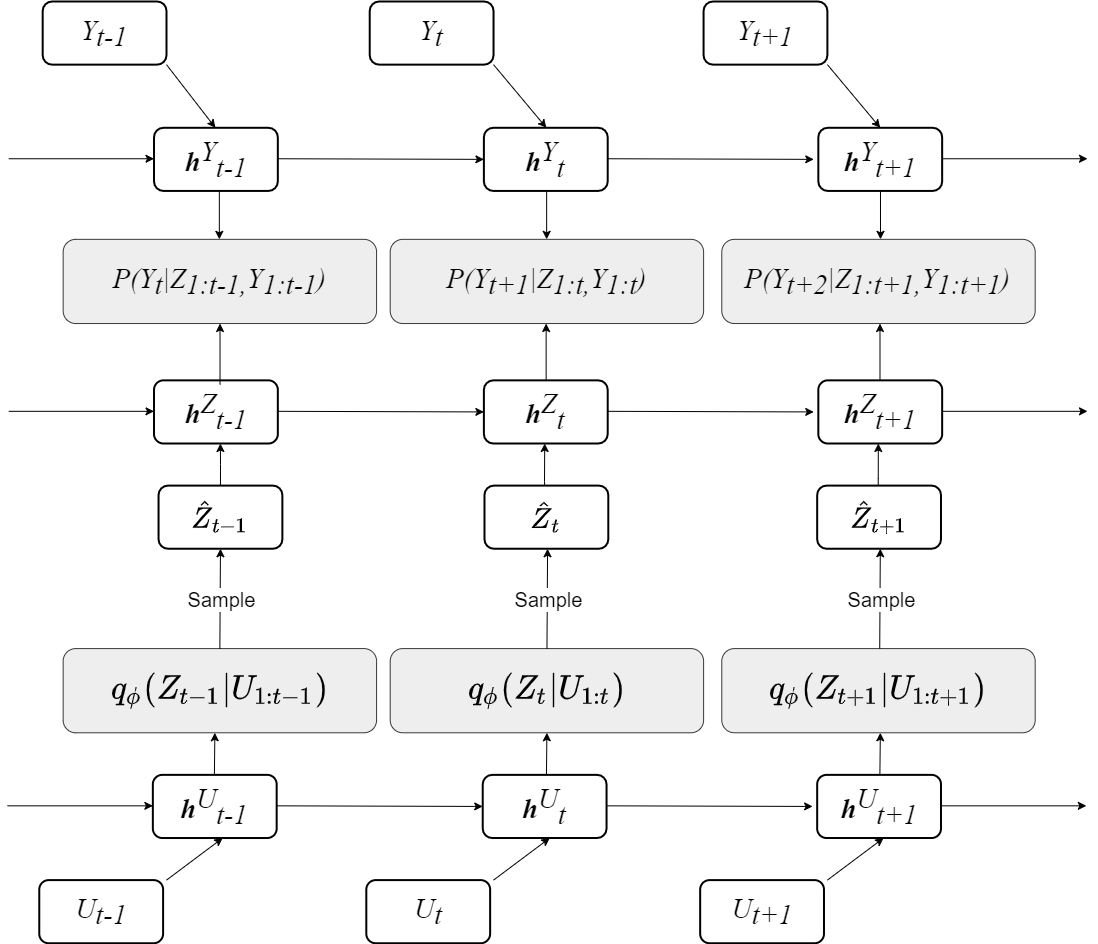}
	\caption{Proposed architecture for the restricted model parameterised by multiple recurrent neural networks.}
	\label{fig2}
\end{figure}

To parameterise the full-model and restricted-model distributions, we leverage the expressive power of recurrent neural networks. In our model we use gated recurrent units (GRU) \citep{Cho2014}. The architecture of the restricted model is given in Fig. \ref{fig2}. Each GRU is characterised by a sequence of hidden states $\boldsymbol{h}^{(i)}_t$ which contains information of time series $i$ up to time $t$. We propose to learn representations of the latent confounder $Z$ using the available proxies $U$ by parameterising the filtering distribution:
\begin{equation}
q_{\phi}(Z_t|U_{1:t})=q_{\phi}(Z_t|\boldsymbol{h}^{(U)}_{t}).
\end{equation}
The inferred representation $\hat{Z}_t$ of $Z_t$ follows an isotropic Gaussian distribution:
\begin{equation}
    \hat{Z}_t\sim q_{\phi}(Z_t|\boldsymbol{h}^{(U)}_{t})= N(\boldsymbol{\mu}(\boldsymbol{h}^{(U)}_{t}),\boldsymbol{\sigma}^2(\boldsymbol{h}^{(U)}_{t})\boldsymbol{I}),
\end{equation}
where the covariance matrix is diagonal. The dimension of $\hat{Z}_t$ is a tunable hyperparameter. The parameters of the filtering distribution are given by
\begin{equation}
(\boldsymbol{\mu},\boldsymbol{\sigma})=f_1(\boldsymbol{h}^{(U)}_{t}),
\label{f1}
\end{equation}
where $f_1$ is a function approximated by an MLP. To ensure positivity of the standard deviation we use a softplus activation function on the MLP output.

\begin{figure}[h]
    \centering
	\includegraphics[width=0.3\columnwidth,clip,keepaspectratio]{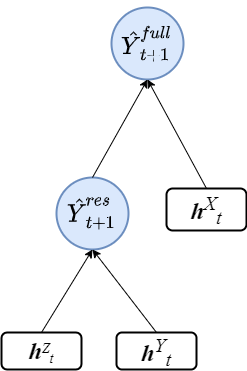}
	\caption{Proposed dual-decoder setup where $\hat{Y}^{res}_{t+1}$ is a prediction sample drawn from the restricted-model distribution $P(Y_{t+1}|Y_{1:t},Z_{1:t})$ shown in Figure \ref{fig2}.}
	\label{fig3}
\end{figure}

To avoid the need to train two separate time series prediction models, we propose the use of a dual-decoder setup. The restricted-model distribution is normal and given as
\begin{equation}
P(Y_{t+1}|Y_{1:t},Z_{1:t})=f_2(\boldsymbol{h}^{(Y)}_{t},\boldsymbol{h}^{(Z)}_{t}).
\label{f2}
\end{equation}
The full-model distribution is also normal and expressed as 
\begin{equation}
P(Y_{t+1}|Y_{1:t},X_{1:t},Z_{1:t})=f_3({\hat{Y}^{res}_{t+1}},\boldsymbol{h}^{(X)}_{t}),  
\label{f3}
\end{equation}
where $\hat{Y}^{res}_{t+1}\sim P(Y_{t+1}|Y_{1:t},Z_{1:t})$ is the predicted value of $Y_t$ from the restricted model and $f_2$ and $f_3$ are two MLP models. The proposed dual-decoder setup is shown in Fig. \ref{fig3} and $\hat{Y}^{full}_{t+1}\sim P(Y_{t+1}|Y_{1:t},X_{1:t},Z_{1:t})$. A combination of Fig.\ref{fig2} and Fig.\ref{fig3} represents the full architecture of the model where the output of the restricted-model serves as one of the inputs of the full-model.

For model optimisation we maximise the following objective function:
\begin{equation}
    L=\sum_{t=1}^{T}\mathbb{E}_{\hat{Z}\sim q_{\phi}}(logP_{\theta_1}(Y_t|Y_{1:t-1},X_{1:t-1},Z_{1:t-1})+logP_{\theta_2}(Y_t|Y_{1:t-1},Z_{1:t-1})),
\end{equation}
where the first and second terms correspond to the full and restricted model distributions respectively, and $\theta_1$ and $\theta_2$ are the model parameters to be optimised.

To infer the Granger causal relationship between $X$ and $Y$ in the presence of a latent confounder, we wish to check whether the inclusion of $X$ in the full-model results in a statistically significant reduction in prediction error compared to the restricted model. With substitute confounders $\hat{Z}_{1:t}$ we perform a two-sample t-test to establish whether $Y_{t+1}\nbigCI X_{1:t}|\hat{Z}_{1:t},Y_{1:t}$ (where $\bigCI$ denotes independence); in such cases we conclude that $X$ Granger-causes $Y$, and vice versa. We use the mean-squared-error $\dfrac{1}{n}\sum_{i=1}^{n}(Y_i-\hat{Y_i})$ as the error metric.

\section{EXPERIMENTS}
\label{experiments}
We first demonstrate the model performance on two arbitrary synthetic datasets with known data generating processes. The nonlinear functions and noise levels have been set arbitrarily. The data generating processes for the two datasets are given by (\ref{dataset1}) and (\ref{dataset2}) respectively. We generate 1000 samples and use 800 for training, 100 for validation and 100 for testing.

\subsection{Dataset 1}
\begin{equation}
\begin{aligned}
Z_t&=tanh(Z_{t-1})+N(0,0.01^2)\\
U_t&=Z^2_{t}+N(0,0.05^2)\\
X_t&=\sigma(Z_{t-2})+N(0,0.01^2)\\
No\ Granger: Y_t&=\sigma(Z_{t-4})+N(0,0.01^2)\\
Granger: Y_t&=\sigma(Z_{t-4})+\sigma(X_{t-2})+N(0,0.01^2)
\end{aligned}
\label{dataset1}
\end{equation}
where the hyperbolic tangent $tanh$ and sigmoid $\sigma$ functions are used to introduce non-linearity into the system. The noise term is Gaussian of the form $N(\mu,std^2)$.
\subsection{Dataset 2}
The data generating processes for $Z$, $U$ and $X$ remain the same as in (\ref{dataset1}). We generate $Y$ using:
\begin{equation}
\begin{aligned}
No\ Granger: Y_t&=Z_{t-3}Z_{t-4}+N(0,0.5^2)\\
Granger: Y_t&=Z_{t-3}Z_{t-4}+X_{t-1}X_{t-2}+N(0,0.5^2)
\end{aligned}
\label{dataset2}
\end{equation}
\subsection{River discharge dataset}
To investigate the model performance on real-world time series, we use the river discharge dataset provided in \cite{Gerhardus2020}. This dataset describes the average daily discharges of rivers in the upper Danube basin. We consider measurements from the Iller at Kempten as $X$, the Danube at Dillingen as $Y$, and the Isar at Lenggries as the proxy variable. All three variables are potentially confounded by rainfall or other weather conditions \citep{Gerhardus2020}. The Iller discharges into the Danube within a day, implying an instantaneous causal link $X\rightarrow Y$. For the scope of Granger causality considered in this paper, the cause is required to precede the effect in time \citep{Eichler2012} so we do not take into account instantaneous causal relationships. We therefore expect no Granger-causal relationship between $X$ and $Y$. The dataset contains roughly 1000 entries, we use 80\% for training, 10\% for validation and 10\% for testing. 

\subsection{Neural network parameters}
The GRU hidden states $\boldsymbol{h}^{(X)}_{t}$,$\boldsymbol{h}^{(Y)}_{t}$ and $\boldsymbol{h}^{(Z)}_{t}$ have a dimension of 5, $\hat{Z}_t$ has a dimension of 1 for the synthetic datasets and 2 for the river discharge dataset, the MLPs $f_1$, $f_2$ and $f_3$ given in (\ref{f1},\ref{f2},\ref{f3}) respectively contain 1 hidden layer with 5 units for the synthetic datasets and 10 units for river discharge, a dropout rate of 0.3 and ReLU activation functions are chosen. We use the ADAM optimiser with a learning rate of 0.001. The sequence length used for model training is 20 with a batch size of 10. These parameters were selected using the validation set through random search. 

\subsection{Statistical testing}
By comparing the sample prediction errors of the full and restricted models, we are able to infer whether a Granger causal relationship between $X$ and $Y$ exists. A two-sample t-test could be used as an additional verification step. For dataset 1\&2 (Granger) we adopt the following null and alternative hypothesis:
\begin{equation}
\begin{aligned}
H_0&: \epsilon_{full}=\epsilon_{restricted}\\ 
H_1&: \epsilon_{full}<\epsilon_{restricted}, 
\end{aligned}
\label{hypo1}
\end{equation}
where $\epsilon_{full}$ and $\epsilon_{restricted}$ are the mean prediction errors generated by the full and restricted models respectively. For dataset 1\&2 (no Granger) we adopt the following alternative hypothesis:
\begin{equation}
H_1: \epsilon_{full}>\epsilon_{restricted}.
\label{hypo2}
\end{equation}
The alternative hypothesis is chosen by comparing the sample mean prediction errors computed by the full and restricted models, I.e. we choose the alternative hypothesis in (\ref{hypo1}) if the mean sample error of the full model is less than that of the restricted model, and vice versa. In cases where the mean sample errors of the two models differ significantly from one another, the statistical test is perhaps redundant. To perform the two-sample t-test, we generate $n=50$ prediction samples from the restricted and full models and we choose a significance level of $\alpha=0.05$. 

\section{RESULTS \& DISCUSSION}
In Table \ref{tab1} we provide the prediction errors of the full and restricted models, the p values of the two-sample t-tests and the Granger causal relationship between $X$ and $Y$ inferred by our model, as well as those inferred by LPCMCI \citep{Gerhardus2020} with $\alpha=0.05$, maximum lag $L=5$ and 4 preliminary iterations, and SVAR-FCI \citep{Malinsky2018} with $\alpha=0.05$ and $L=5$. These are conditional independence based methods for inferring potential causal relationships and are capable of handling latent confounders. 

\begin{table}[h]
    \centering
	\caption{Table showing the prediction errors of the full and restricted models, p values of two-sample t-tests and the inferred Granger causal relationship given by our model, LPCMCI and SVAR-FCI. The symbol $\times$ denotes that the model finds a Granger non-causal relationship between $X$ and $Y$.}
	\resizebox{\textwidth}{!}{
	\begin{tabular}{|c|c|c|c|c|c|c|}
		\hline
		Dataset & Restricted-model error ($mean\pm std$) & Full-model error ($mean\pm std$) & p value & Ours & LPCMCI & SVAR-FCI\\
		\hline
		dataset 1 (Granger) & $4.99\times10^{-2}\pm 6.00\times10^{-4}$ & $1.76\times10^{-2}\pm 5.71\times10^{-5}$ & $<1.00\times10^{-3}$ & \checkmark & \checkmark & \checkmark\\
		\hline
		dataset 1 (no Granger) & $2.03\times10^{-2}\pm 4.00\times10^{-4}$ & $3.54\pm 2.00\times10^{-4}$ & $<1.00\times10^{-3}$ & $\times$ & \checkmark & \checkmark\\
		\hline
		dataset 2 (Granger) & $2.07\times10^{-1}\pm 7.00\times10^{-4}$ & $2.03\times10^{-1}\pm 9.75\times10^{-5}$ & $<1.00\times10^{-3}$ & \checkmark & $\times$ & $\times$\\
		\hline
		dataset 2 (no Granger) & $1.56\times10^{-1}\pm 1.85\times10^{-4}$ & $1.60\times10^{-1}\pm 5.73\times10^{-6}$ & $<1.00\times10^{-3}$ & $\times$ & $\times$ & $\times$\\
		\hline
		river discharge & $4.85\times10^{-2}\pm1.50\times10^{-3}$&$6.10\times10^{-2}\pm1.12\times10^{-3}$& $<1.00\times10^{-3}$& $\times$ &$\times$ &$\times$\\
		\hline
	\end{tabular}}
	\label{tab1}
\end{table}

We observe from Table \ref{tab1} that the $p\  value<0.05$ for all the statistical tests. For dataset 1\&2 (no Granger) and the river discharge dataset, we reject the null hypothesis that the mean prediction errors of the restricted and full models are equal and conclude that the inclusion of $X$ to predict future values of $Y$ results in a higher prediction error and therefore $X$ does not Granger-cause $Y$. For dataset 1\&2 (Granger) we reject the null hypothesis and conclude that the inclusion of $X$ reduces the prediction errors of $Y$ and therefore $X$ Granger-causes $Y$. Our model correctly identifies the correct Granger-causal relationship in all scenarios, whereas LPCMCI and SVAR-FCI identify spurious relationships for dataset 1 (no Granger) and dataset 2 (Granger).

Real-world time series can be highly nonlinear and have different noise levels. We have shown that our model is able to identify the Granger-causal relationship for various nonlinear functions and arbitrary noise levels. We investigate the robustness of our model by varying the signal-to-noise ratio defined as:
\begin{equation}
\gamma=\frac{\frac{1}{T}\sum_{t=1}^{T}|s_t|}{\sigma},
\end{equation}
where $|s_t|$ denotes the magnitude of the signal ($Y_t$ without the noise term) at $t$ and $\sigma$ is the standard deviation of the noise term in the data generating process. For dataset 1\&2 (Granger) we wish to find the critical $\gamma$ below which the noise term becomes dominant and the model fails to identify the Granger-causal link between $X$ and $Y$; to do this we vary the standard deviation $\sigma$ of the noise term in (\ref{dataset1}) and (\ref{dataset2}). We start with a rough range of $\gamma=10$ to $\gamma=100$ and we use a bisection search strategy to find the critical value $\gamma^*$. A $p\ value<0.05$ denotes Granger causality inferred by our model. Results are shown in Table \ref{tab2}. For dataset 1 (Granger) we see that the critical value $\gamma^*$ is approximately 59.22 (highlighted in bold), i.e. the Granger-causal link between $X$ and $Y$ for this set of stochastic time series can only be identified if $\gamma\ge59.22$; for dataset 2(Granger) $\gamma^*\approx27.58$. 

Lastly, we tested the sensitivity of the model output to the sequence length $\tau \in \{4,6,8,10,12,14,16\}$ used in training for dataset 1\&2 (Granger). We noted that all $p\ value<0.001$, which suggests that our model is able to consistently identify the Granger-causal link given short and long $\tau$ used in training. This is desirable as it indicates that model results are not very sensitive to the choice of hyperparameters. 

\begin{table}[h]
    \centering
	\caption{Sensitivity analysis of model performance with varying signal-to-noise ratio $\gamma$.}
	\resizebox{0.5\textwidth}{!}{
	\begin{tabular}{|c|c|c|c|}
		\hline
		Dataset 1 (Granger) $\gamma$ & $p\ value$ & Dataset 2 (Granger) $\gamma$ & $p\ value$\\
		\hline
		10.00 & 1.00 & 10.00 & 1.00\\
		\hline
		55.00 & $9.99\times10^{-1}$ & 21.25 & $9.99\times10^{-1}$\\
		\hline
		57.81 & $9.41\times10^{-1}$ & 26.88 & $5.44\times10^{-2}$\\
		\hline
		58.51 & $4.10\times10^{-1}$ & \textbf{27.58} & \textbf{$4.73\times10^{-3}$}\\
		\hline
		\textbf{59.22} & \textbf{$1.93\times10^{-2}$} & 28.28 & $<1.00\times10^{-3}$\\
		\hline
		60.63 & $<1.00\times10^{-3}$ & 29.69 & $<1.00\times10^{-3}$\\
		\hline
		66.25 & $<1.00\times10^{-3}$ & 32.50 & $<1.00\times10^{-3}$\\
		\hline
		77.50 & $<1.00\times10^{-3}$ & 55.00 & $<1.00\times10^{-3}$\\
		\hline
		100 & $<1.00\times10^{-3}$ & 100 & $<1.00\times10^{-3}$\\
		\hline
	\end{tabular}}
	\label{tab2}
\end{table}

\section{CONCLUSION}
In this paper we have presented a deep-learning based approach to model nonlinear Granger-causality with in the presence of a latent confounder. Our model involves the use of multiple recurrent neural networks to parameterise a restricted-model distribution $P(Y_{t+1}|Y_{1:t},Z_{1:t})$ and a full-model distribution $P(Y_{t+1}|Y_{1:t},X_{1:t},Z_{1:t})$. We generate prediction samples from the two distributions and we use a two-sample t-test to establish whether the inclusion of $X$ helps to predict future values of $Y$ given a learned representation of the confounder. To enable efficient comparison, we propose a dual-decoder setup, which avoids the need to train two separate models (as presented in many existing literature), and we believe this helps to reduce bias resulting from neural network hyperparameter tuning. We demonstrate the effectiveness of our model on both synthetic and real-world datasets, and we recognise that a high enough signal-to-noise ratio is required to correctly identify a Granger-causal link.

\section*{Acknowledgements}
The authors acknowledge Dr Fabio Caccioli and Dr Brooks Paige (both Dept of Computer Science, UCL) for their advice on the project and the manuscript.
\bibliography{sample}

\end{document}